\pdfoutput=1

\documentclass[11pt]{article}
\usepackage{multirow}

\usepackage{ACL2023}

\usepackage{times}
\usepackage{latexsym}
\usepackage[rightcaption]{sidecap}
\usepackage[section]{placeins}
\usepackage{float}
\usepackage{caption}
\usepackage{subcaption}
\usepackage[export]{adjustbox}

\usepackage{graphicx} 
\graphicspath{ {figures/} }

\usepackage[T1]{fontenc}

\usepackage[utf8]{inputenc}

\usepackage{microtype}

\usepackage{inconsolata}

\title{Comparing Data Augmentation Methods for End-to-End Task-Oriented Dialog Systems}

\author{Christos Vlachos$^{1,2}$, Themos Stafylakis$^{1,2,3}$, Ion Androutsopoulos$^{1,3}$\\
	\normalsize $^{1}$Department of Informatics, Athens University of Economics and Business, Greece\\
	\normalsize $^{2}$Omilia Natural Language Solutions Ltd.
    \normalsize $^{3}$Archimedes/Athena RC, Greece
}

\begin{document}
\maketitle
\begin{abstract}
Creating effective and reliable task-oriented dialog systems (ToDSs) is challenging, not only because of the complex structure of these systems, but also due to the scarcity of training data, especially when several modules need to be trained separately, each one with its own input/output training examples. Data augmentation (DA), whereby synthetic training examples are added to the training data, has been successful in other NLP systems, but has not been explored as extensively in ToDSs. We empirically evaluate the effectiveness of DA methods in an end-to-end ToDS setting, where a single system is trained to handle all processing stages, from user inputs to system outputs. We experiment with two ToDSs (UBAR, GALAXY) on two datasets (MultiWOZ, KVRET). 
We consider three types of DA methods (word-level, sentence-level, dialog-level), comparing eight DA methods that have shown promising results in ToDSs and other NLP systems. We show that all DA methods considered are beneficial, and we highlight the best ones, also providing advice to practitioners. 
We also introduce a more challenging few-shot cross-domain ToDS setting, reaching similar conclusions.  

\end{abstract}


\section{Introduction} \label{sec:introduction}
Task-oriented dialog systems (ToDSs) aim to understand and fulfil a user's goal in a particular application domain (e.g., booking tickets, restaurant reservations) through natural language conversation \citep{2,3}. They traditionally include several modules to handle a variety of processing stages. A natural language understanding (NLU) module typically extracts structured information from each user's utterance, usually in the form of slot-value pairs (e.g., for destination, travel date). A dialog state tracker (DST) keeps track of the information that has been revealed about the user's goal (dialog state) throughout the conversation. A policy optimizer (PO) decides the system's next action (e.g., request information for a particular unfilled slot of the dialog state, or recommend a restaurant). A natural language generation (NLG) module realises communication actions of the system in natural language. In spoken conversations, automatic speech recognition (ASR) and text-to-speech (TTS) modules are also included \citep{58}, though here we consider only written conversations.

When the modules of a ToDS are trained separately, large volumes of annotated training data are needed, since each module requires its own input/output training examples. In end-to-end ToDSs \citep{6,78}, a single system is jointly trained to handle all (or most) of the processing stages, reducing (or at least harmonising) the kinds of annotated training data required, since each training example can be used by all (or most) of the modules \cite{8}. Recent Transformer-based end-to-end ToDSs \citep{1,5,20} aim to reduce the necessary training data further, by leveraging pre-training on unlabelled data.  
However, the best performance is still obtained by fine-tuning on domain-specific annotated training datasets \cite{59}, which are costly to obtain, especially when moving to a new domain (e.g., from booking airline tickets to reserving restaurants). LLMs that promise adequate results in few- or even zero-shot settings \citep{14,41} have been under-explored in ToDSs, and  work along this direction has focused mostly on the DST module, with contradicting results \cite{59, 60}. Thus cost-effective methods to obtain or augment training sets for ToDSs are needed.
 
We experiment with data augmentation (DA), which adds synthetic training examples to an existing annotated (possibly small) training set. DA methods have shown promising results in various NLP tasks, e.g., text classification, text generation, question answering \citep{61,62,21}, but have not been systematically compared in end-to-end ToDSs (Section~\ref{sec:related}). We simulate the case where the available annotated training set is limited, which is often the case with real-life ToDS applications, especially when moving to new application domains. We experiment with two end-to-end pre-trained Transformer-based TODSs, namely UBAR \citep{17} and GALAXY \citep{81}, and two English textual conversational datasets, MultiWOZ 2.0 \citep{10} and KVRET \citep{82}. We vary both the size of the initial annotated training set (e.g., 2\%, 10\%, 25\% of MultiWOZ's training set) and the expansion factor (generating one, two, or four synthetic examples from each original one, i.e., x2, x3, x5 expansion). We also introduce a more challenging \emph{few-shot cross-domain setting}: in each iteration, we select a single domain as the target one, we remove from the training set all but 20 (few-shot) dialogs of the target domain, we generate synthetic training dialogs from the 20 dialogs, and evaluate only on test dialogs from the left-out target domain.\footnote{For example, the domains of MultiWOZ are: \textit{train}, \textit{taxi}, \textit{hotel}, \textit{restaurant}, \textit{hospital}, \textit{police} and \textit{attraction}.}

We compare eight DA methods that have shown promising results in ToDSs (mostly when training individual modules) 
and other NLP systems. The DA methods we consider are of three types. \emph{Word-level} augmentation methods replace words of the original training examples by similar words, e.g., words that have similar word embeddings or words suggested by models that predict masked words.

\emph{Sentence-level} augmentation includes paraphrasing sentences of the original training data via back-translation (translating to another language and back), invoking systems trained specifically to perform paraphrasing, or by prompting Large Language Models (LLM); we also include in this type methods that rotate (swap) parts of the dependency tree of the original sentence. Finally, \emph{dialog-level} augmentation methods exploit dialog-specific annotations of the original training examples (e.g., dialog states), which are available in ToDS datasets.

Our contributions are: (1) We conduct a systematic comparison of DA methods in an end-to-end ToDS setting, the largest (in terms of DA methods) comparison to date in this setting. (2) We show that substantial performance gains can be obtained with DA methods, even when using pre-trained models, and we offer concrete advice to ToDS practitioners (Section~\ref{sec:advice}). (3) We introduce and utilise a new few-shot cross-domain evaluation setting for DA approaches in ToDS, where we show that DA methods again boost performance. 

\section{Data Augmentation Methods} \label{sec:body} \label{sec:da_methods}
\subsection{Word-level Augmentation}
\label{sec:3_1_subsection} \label{sec:word_level_augmentation}
Word-level DA methods replace words of the original training examples by similar words, e.g., words with similar word embeddings or words suggested by models that predict masked words \citep{47, 51}. We experiment with word substitutions using either Word2Vec \citep{43} or RoBERTa \citep{44}. Apart from boosting performance, word-level DA can generate more synthetic instances per original one, compared to other DA types considered.
 
\subsubsection{Word2Vec-based Word Substitution} \label{sec:3_1_1_subsection}
We employ Google's 300-dimensional embeddings, created using Word2Vec.\footnote{Available via Gensim  \citep{46}.} We exclude stop-words or delexicalised tokens (special tokens replacing slot values) as candidates for DA. For every word in each dialog utterance, we find the 10 most similar ones in terms of cosine similarity of word embeddings. Preliminary testing showed that this may actually deteriorate performance, as some word replacements create dialog states and system actions inconsistent with the original ones, in agreement with \citet{27}, who note that swapping certain words may change the original semantics. Instead, for every experiment, we train the ToDS with the original training data (the percentage available per experiment). For every word position, we test every candidate substitution from the pool of 10 most similar ones selecting at random, by passing the augmented sentence to that model. If the model's prediction (in terms of dialog state and system action) is different than the original one, the augmentation is discarded and the next word replacement from the pool is considered. 

\subsubsection{LM-based Word Substitution} \label{sec:3_1_2_subsection}
Masking a word and using a language model (LM) to predict the masked token is a more advanced case of word substitution, previously used in numerous cases \citep{48, 49, 27}. We leverage the large version of an out-of-the-box RoBERTa \citep{44}. The list of candidate substitutions now comprises words that the LM deemed more likely, otherwise the method remains as in Word2Vec-based substitution, including the requirement that the dialog state and system action must remain the same. 

\subsection{Sentence-level Augmentation} \label{sec:3_2_subsection}
Sentence-level augmentation includes paraphrasing (via back-translation, with systems trained specifically to perform paraphrasing, or by prompting LLMs) and fragment rotation (which moves parts of the dependency tree). These methods cannot generate as many synthetic training instances as word-level substitution, but their synthetic utterances are semantically closer to the original ones.

\subsubsection{Back-translation} \label{sec:3_2_1_subsection}
Back-translation has been used in many NLP tasks \citep{30,56,57}. It translates a piece of text (an utterance in our case) to a different (pivot) language and back to the original one. We use French as the pivot language and the Google translate API.\footnote{More languages could be used to generate more data.} As explained by \citet{32}, it is important to retain the original names of entities, such as attraction names and addresses. Special tokens used for delexicalisation also face the same issue. To avoid wrongly back-translating these special tokens, we apply back-translation to each utterance only after delexicalising and replacing the special tokens with special numbers (preceded by '\#'), which remain unchanged during translation. The special tokens are relexicalized after DA.

\subsubsection{PEGASUS Paraphrasing} \label{sec:3_2_2_subsection}
The utilisation of LMs for paraphrasing was beneficial in the ToDS work of \citet{22}, \citet{24}, and \citet{26}. Following \citet{33}, we leverage an instance of the Transformer-based PEGASUS \citep{35}, fine-tuned for paraphrasing.\footnote{\url{huggingface.co/tuner007/PEGASUS\_paraphrase}} 

We set PEGASUS to generate up to two paraphrased utterances, and randomly select one, an approach that allows us the generate multiple synthetic dialogs per training instance.\footnote{More paraphrases can be generated, but most of them tend to be of lower quality and may lead to wrong dialog states.} 
For each synthetic utterance, we check whether the delexicalisable tokens are preserved. If not, the augmentation is not applied. 

\subsubsection{Fragment Rotation} \label{sec:3_2_3_subsection}
Cropping or rotating/flipping images is a popular label preserving DA approach in Computer Vision \citep{37, 38}. An NLP analog of rotation was proposed by \citet{36} and has also been tested in a ToDS setting \citep{71, 23}. The main idea is to rotate (swap) in the dependency tree of a sentence, fragments (such as subjects/objects) around the root (usually a verb). We leverage Stanford's dependency parser \citep{40} to generate annotations for MultiWOZ 2.0, consistent with the Universal Dependencies 2.1 \citep{39} in \c{S}ahin and Steedman's work. We then employ their method to generate synthetic utterances.\footnote{\url{https://github.com/gozdesahin/crop-rotate-augment}} This may generate multiple outputs per original training instance, from which we sample uniformly. This method is a better fit for more  free word-order languages, but both in \citet{71} and in our experiments, it leads to substantial improvements in English too.

\subsubsection{LLM Paraphrasing} \label{sec:3_2_4_subsection}
LLMs have shown impressive results in tasks they have not encountered during training \citep{41,42}. Here we leverage ChatGPT, GPT-3.5-turbo specifically, and prompt it to generate augmented data via paraphrasing. We limit ourselves to two synthetic utterances to minimise the LLM's usage cost, randomly selecting one of the two, provided it includes all of the tokens that will be delexicalised. As a prompt we use a simple instruction and no in-context learning.\footnote{More information can be found in Appendix \ref{sec:appendix}.}

\subsection{Dialog-level Augmentation} \label{sec:3_3_subsection}
Dialog-level DA methods exploit dialog-specific annotations that ToD datasets include (e.g., dialog states, system actions). These methods can generate more synthetic data than sentence-level DA methods, but fewer than word-level ones. Also, the dialog-level DA methods considered here assume that dialog turns (user utterance--system response pairs) that have the same dialog state convey the same information, which is not always the case; hence, they may produce erroneous data. 

\subsubsection{Dialog Tree}\label{sec:3_3_1_subsection} 
The first dialog-level DA method we test was originally proposed by \citet{11}. Based on
 their approach, we create a tree structure containing dialogs that can be formed by combining the turns of the available training data in three steps. The first step is the creation of turn templates. Each template comprises a delexicalised turn, along with the current delexicalised dialog state ($\textit{Cds}$) and the delexicalised dialog states of the previous ($\textit{Pds}$) and next ($\textit{Nds}$) turns of the dialog they originate from. If a template refers to the first dialog turn, its $\textit{Pds}$ is ``Root''. If a template refers to the last turn, its $\textit{Nds}$ is ``Leaf''. Let the set of templates created in the first step be \( T = \{r, t_1, ..., t_n\} \), where $r$ is a special root template used in the second step. 

In the second step, templates are linked in a tree structure under a certain condition. A template $t_j$ can be linked with a template $t_i$ as its child node, only if $t_i$'s $\textit{Nds}$ matches $t_j$'s $\textit{Cds}$ and $t_j$'s $\textit{Pds}$ matches $t_i$'s $\textit{Cds}$ (see Fig. \ref{fig:d_tree} of Appendix \ref{sec:appendix_c} for an example). All templates $t$ that appear at the beginning of a dialog are automatically assigned as children of the special root template $r$. The third step is surface realisation, where delexicalised slots are filled with slot values.

As the time complexity of this method increases exponentially in the number of dialogs, we sample only 50 dialogs from the original training set in the first step. We then create the tree and generate a new synthetic dialog by starting from the root and randomly selecting a child recursively until we reach a leaf. We then repeat, sampling again 50 dialogs (with replacement) from the original training set and generating another synthetic dialog.

\subsubsection{Act-Response Substitution} \label{sec:3_3_2_subsection}
Multi-Action Data Augmentation (MADA) is another DA approach based on dialog state matching, proposed by \citet{12}. For every turn, they identify all other turns (of the original training dialogs) that have the same delexicalised dialog state and extract the system actions associated with the identified turns. They then train their ToDS using the original and extracted system actions per turn. Normally, a ToDS would encounter a single system action in each instance and \citet{12} observe that some system actions are more likely to be the ground truth actions, compared to others that may be correct as well. Instead, through MADA, the model encounters a more balanced set of system actions, helping it generate diverse responses during inference. In our case, along with valid system actions, we also store the corresponding system responses per turn, following the same principle. Then, for each dialog turn we replace the original system action and response by sampling from the set created by the process discussed.


\section{Experimental Setup} \label{sec:setup}

In all experiments, we use three different randomly sampled training subsets (e.g., 3 different 10\% subsets) and report averages.

\subsection{Main Experiments} \label{sec:main_setup}
\textbf{MultiWOZ:} For our main experiments, we use MultiWOZ, an English human-to-human conversational dataset produced by a Wizard-of-Oz experiment, which has been used extensively in previous work \citep{52, 53, 54, 7, 55}. MultiWOZ 2.0, the most commonly used version that we also use, includes dialogs spanning 7 domains. It includes 10,438 dialogs in total, of which 1,000 are reserved for validation and another 1,000 for testing. Most dialogs fall into two or more domains making it a challenging benchmark. 
 Of the tasks supported by MultiWOZ 2.0 (e.g., DST, intent classification, action-to-response generation) we select the response generation (RG) task, which UBAR and GALAXY can handle end-to-end.

\smallskip\noindent\textbf{Evaluation measures:} Following \citet{10}, we mainly report \textit{Score} values for MutiWOZ. We also present \textit{Inform}, \textit{Success}, and \textit{BLEU} results in the appendix. \textit{Inform} is the percentage of dialogs where the system provided a correct recommendation based on the user's constraints (e.g., cheap hotel). \textit{Success} is the percentage of dialogs where the system satisfied the user's constraints ($\textit{Inform} = 100\%$) and provided all the requested information \citep{10,19}. \textit{BLEU} \citep{77} measures the word $n$-gram similarity between the system and the gold response.
Finally, $\textit{Score} = \frac{1}{2}(\textit{Inform} + \textit{Success}) + \textit{BLEU}$.

\smallskip\noindent\textbf{UBAR:}
For the MultiWOZ experiments, we use UBAR \citep{17}, a GPT2-based ToDS \citep{14} that remains competitive as shown in recent studies \citep{19, 20}, and is also easier (and much less expensive) to use compared to larger LLMs. UBAR is also based on the framework used by MADA \citep{12}, thus facilitating experimenting with dialog-level DA methods.

\smallskip\noindent\textbf{MultiWOZ with all DA methods:} In a first set of experiments, we reduce the size of the MultiWOZ training set to 850 dialogs ($\sim$10\% of the training set) and produce an extra synthetic dialog from each original one, doubling ($\times 2$ ) the training set. 

\smallskip\noindent\textbf{MultiWOZ with best three DA methods:}
In a second set of experiments, we consider only the best three DA methods of the previous experiments, but we now consider generating one, two, or four synthetic dialogs from each original dialog  ($\times 2$, $\times 3$, $\times 5$), using only $\sim$2\% (170), $\sim$10\% (850), or $\sim$25\% (2,125) of the MultiWOZ training dialogs.

\smallskip\noindent\textbf{Few-shot cross-domain experiments:}
Finally, we test the best three DA methods of the first set of experiments, now in a few-shot, cross-domain setting for the \textit{restaurant}, \textit{hotel} and \textit{attraction} domains of MultiWOZ. As in the first set of experiments, we use 10\% of the MultiWOZ training dialogs. Additionally, in three iterations, we leave one domain out of the training 
set apart from 20 (few-shot) dialogs, and we use only test (and validation) dialogs that include the left-out domain. 
We generate four synthetic dialogs for each few-shot dialog (x5) in these experiments, which was the augmentation size that worked best in the second set of experiments when there were very few training instances.

\subsection{Additional Experiments}\label{sec:secondary_setup}

To ensure that our previous findings are not dataset- and model-specific, we repeat them using KVRET  \citep{82} and GALAXY \citep{81}.

\smallskip\noindent\textbf{KVRET} is also an English multi-domain dataset, the domains being \textit{weather}, \textit{navigation}, \textit{schedule}. It includes 2,424 dialogs, with 302 being reserved for validation and 304 for testing. Contrary to MultiWOZ, each dialog belongs in a single domain. 

\smallskip\noindent\textbf{Evaluation measures:} Following \citet{81}, for KVRET we report \textit{Score} (calculated using the same formula as in MultiWOZ), with slight modifications. \textit{Match} is reported, which is equivalent to the \textit{Inform} of MultiWOZ. Moreover, \citet{78} argues that the original \textit{Success} rate favours recall, as a dialog would be considered successful if the model responded with every possible recommendation. Hence \textit{Success F1} is used to balance precision and recall instead of \textit{Success}.
 
\smallskip\noindent\textbf{GALAXY} is a more recent ToDS, pretrained using UniLM \citep{83}, 
currently among the best 
ToDSs on KVRET \cite{86}.

\smallskip\noindent\textbf{Experiments:} We conduct similar experiments on KVRET as on MultiWOZ. The initial training sets (before DA) include $\sim$5\% (120), $\sim$10\% (250), $\sim$25\% (600) dialogs from KVRET's training set.
We select the three best performing DA approaches on the 5\% subset and conduct the same experiments using the 10\% and 25\% subsets, generating one, two, or four synthetic instances from each original one (x2, x3, x5). KVRET has an average dialog length of ~2.5 turns, contrary to MultiWOZ's 13.68. 
We exclude the two dialog-level DA methods from the KVRET experiments, since they had the worst performance in the first set of experiments on MultiWOZ.
As each dialog belongs in a single domain, our cross-domain evaluation setting is also not applicable to KVRET.

\section{Experimental Results and Analysis} \label{sec:result}

\subsection{Results of the Main Experiments}

\textbf{MultiWOZ with all DA methods:} The results of the first set of experiments (all 8 DA methods, 10\% of MultiWOZ's training set, x2 expansion) are presented in Table~\ref{tab:10_x2_augment_small}. All DA methods lead to a substantial performance boost. The three highest scores are obtained with Word2Vec substitution, PEGASUS paraphrasing and fragment rotation, 
improving UBAR without DA (``10\% training set'') by about 9 points. Surprisingly, fragment rotation is among the best DA method, even though English is not particularly free word-order (Section \ref{sec:3_2_3_subsection}). 
The worst results come from the dialog-level DA methods, presumably because matching delexicalised dialog states introduces noise \cite{11}.

\begin{figure}[!ht]
\begin{center}
\includegraphics[scale=0.5]{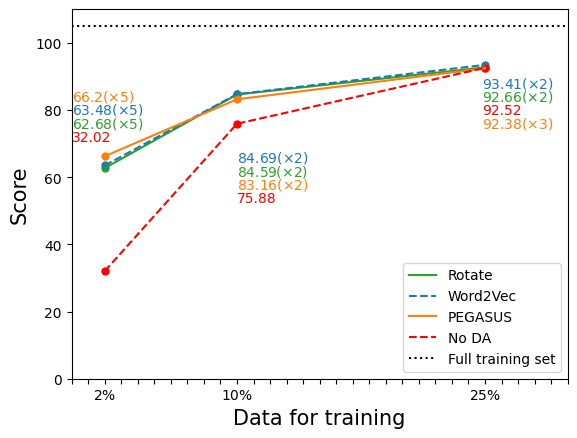} 
\vspace{-3mm}
\caption{\textbf{MultiWOZ Score}, for the \textbf{3 best DA methods} of Table~\ref{tab:10_x2_augment_small}, using \textbf{2\%, 10\%, 25\%} of the original training dialogs. We report in brackets the expansion size (\textbf{x2, x3, x5}) that led to the best \textit{Score}.}
\label{figures/all_WOZ.png}
\vspace{-5mm}
\end{center}
\end{figure}

\smallskip\noindent\textbf{MultiWOZ with best three DA methods:} The results of the second set of experiments (best 3 DA methods, 2\%, 10\%, 25\% of original training set, expansion by x2, x3, x5) are shown in Fig. \ref{figures/all_WOZ.png}. For each original training set size (horizontal axis), we show the \textit{Score} (vertical axis) of each DA method (curves). To save space, we show scores using the best (per DA method and training set size) expansion size (x2, x3, x5). As expected, DA methods are more beneficial when the original annotated training set is limited (e.g. 2\%, 10\%), with much smaller improvements when more annotated data are available (25\%).
Also, when the original training set is very limited (2\%), x5 expansion leads to the best results, but when the original training set is larger (10\%, 25\%), more conservative (x3 and  x2, respectively) expansion is better, presumably because it becomes more important to avoid adding multiple variants of the same original training instance. Overall, the three best DA methods lead to very similar performance boosts, in agreement with the results of the previous experiments (Table~\ref{tab:10_x2_augment_small}).

\begin{figure}[!ht]
\begin{center}
\includegraphics[scale=0.45]{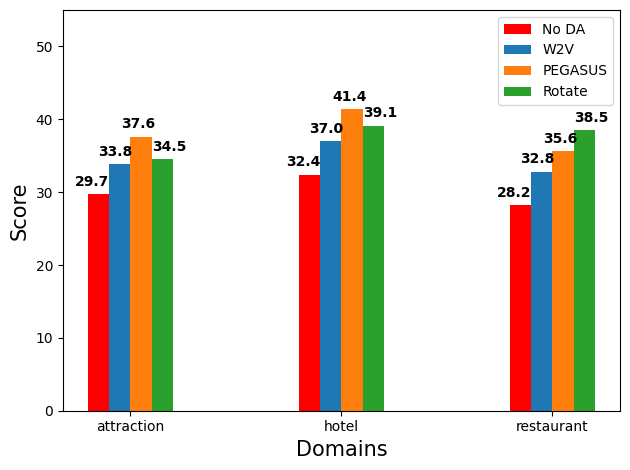} 
\vspace{-3mm}
\caption{\textbf{MultiWOZ Score}, for the \textbf{3 best DA methods} of Table~\ref{tab:10_x2_augment_small}, in the \textbf{few-shot cross-domain} setting (domains: \textit{attraction}, \textit{hotel}, \textit{restaurant}). 
}
\label{fig:expect_domain}
\vspace{-5mm}
\end{center}
\end{figure}

\smallskip\noindent\textbf{Few-shot cross-domain experiments:}
The results of these experiments are shown in Fig.~\ref{fig:expect_domain}. All DA methods led to substantial improvements in all three few-shot domains, increasing the \textit{Score} by up to 8, 9 and 10 points for the \textit{attraction}, \textit{hotel} and \textit{restaurant} domains, respectively. Contrary to previous experiments, here PEGASUS paraphrasing was more effective overall compared to the other two DA methods.\footnote{More detailed results of the main and additional experiments can be found in Appendix \ref{sec:appendix_b}.}

\begin{table}[h!]
\begin{center}
\begin{tabular}{lll}
\hline
\textbf{DA Type} & \textbf{DA Method} & \textbf{Score} \\
\hline
\multirow{2}{4em}{No DA} & Full training set & 105.1 \\& 10\% training set & 75.88\\ 
  \hline
\multirow{2}{6em}{Word-level} & Word2Vec replace &  84.69\\ 
     & RoBERTa replace & 80.11\\ 
  \hline
  \multirow{4}{6em}{Sentence-level} & Back-translation (FR) & 81.31\\   
   & PEGASUS  & 83.16\\   
   & Fragment rotation & 84.59\\
  & LLM paraphrase & 81.48\\
  \hline
  \multirow{2}{6em}{Dialog-level} & Dialog Tree & 79.94\\
  & Act-Response & 80.78\\
  \hline
\end{tabular}
\caption{\textbf{MultiWOZ Score}, for all \textbf{8 DA methods}, using \textbf{10\%} of the original training dialogs,  producing a single synthetic dialog from each original one (\textbf{x2}).
}
\label{tab:10_x2_augment_small}
\vspace{-5mm}
\end{center}
\end{table}

\subsection{Results of the Additional Experiments}

\textbf{KVRET with all DA methods:} GALAXY's results with all 6 applicable (to KVRET) DA methods, using 5\% of KVRET's training dialogs, and x2 expansion are reported in Table~\ref{tab:10_x2_augment_small_kvret}. Word2Vec substitution and fragment rotation remain among the top performers, with the LLM-based paraphrasing marginally overtaking PEGASUS paraphrasing. 

\begin{figure}[!ht]
\begin{center}
\includegraphics[scale=0.5]{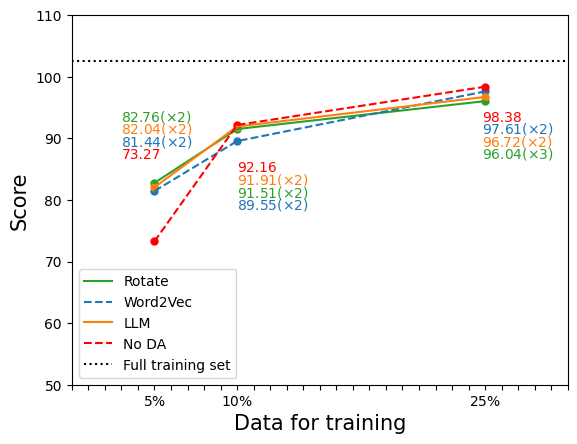} 
\vspace{-3mm}
\caption{\textbf{KVRET Score}, for the \textbf{3  best DA methods} of Table~\ref{tab:10_x2_augment_small_kvret}, using \textbf{5\%, 10\%, 25\%} of the original training dialogs. We report in brackets the expansion size (\textbf{x2, x3, x5}) that led to the best \textit{Score}.}
\label{figures/all_KVRET.png}
\vspace{-5mm}
\end{center}
\end{figure}

\smallskip\noindent\textbf{KVRET with best three DA methods:} Figure~\ref{figures/all_KVRET.png} shows results obtained with the best three DA methods of the previous experiment, now using 5\%, 10\%, 25\% of KVRET's training set, and x2, x3, x5 expansion. Interestingly, DA (any method) manages to improve performance only when 5\% of KVRET's training set is used.\footnote{This is also why we select the best three DA methods using 5\% of KVRET's training set in the previous experiment.}
Although KVRET is smaller than MultiWOZ (2,424 dialogs in total vs.\ 10,438) and one would expect DA methods to be more useful in KVRET, it is also much easier than MultiWOZ, because each dialog concerns a single domain. This is evident by the fact that GALAXY with only 25\% of the KVRET training set and no DA (Fig.~\ref{figures/all_KVRET.png}, rightmost result of the red dashed line) is almost as good as GALAXY trained on the entire KVRET training set, again with no DA (horizontal dotted line). Even with only 10\% of KVRET's training set, GALAXY with no DA (red dashed line) obtains a \textit{Score} of 92.16, leaving very little scope for improvement. 
When using 5\% of the original training set, where there is much larger scope for improvement (Fig.~\ref{figures/all_KVRET.png}, left), all DA methods substantially improve performance, and the best DA method (fragment rotation) improves performance by approx.\ 9 points (Table~\ref{tab:10_x2_augment_small_kvret}). The results of Figures~\ref{figures/all_WOZ.png} and \ref{figures/all_KVRET.png} also indicate that there is no universal threshold of initial annotated training instances, below which DA is beneficial; the threshold depends on the difficulty of the particular task (e.g., whether dialogs are single- or multi-domain). Instead, one should examine performance; when it is very low (e.g., with 2\% of MultiWOZ or 5\% of KVRET), DA methods are definitely worth considering, otherwise when performance is already very high (e.g., exceeding 90 \textit{Score} when using 10\% or 25\% of KVRET), applying DA is probably pointless.
In all cases, generating more than one or two synthetic examples per original one (more than x2 or x3 expansion) is not beneficial.

\begin{table}[h!]
\begin{center}
\begin{tabular}{lll}
\hline
\textbf{DA Type} & \textbf{DA Method} & \textbf{Score} \\

\hline
\multirow{2}{4em}{No DA} & Full training data & 102.5 \\& 5\% training data &  73.27\\ 
  \hline
\multirow{2}{6em}{Word-level} & Word2Vec replace &  81.44\\ 
     & RoBERTa replace &  80.67\\ 
  \hline
  \multirow{4}{6em}{Sentence-level} & Back-translation (FR) & 77.83\\   
   & PEGASUS  & 75.78\\   
   & Fragment rotation & 82.76\\
  & LLM paraphrase & 82.04\\
  \hline
\end{tabular}
\caption{\textbf{KVRET Score}, for \textbf{6 applicable DA methods}, using \textbf{5\%} of original training dialogs, producing a single synthetic dialog from each original one (\textbf{x2}).}
\vspace{-5mm}
\label{tab:10_x2_augment_small_kvret}
\end{center}
\end{table}

\subsection{Response and Error Analysis}\label{sec:analysis}
\begin{figure}[!ht]
\begin{center}
\includegraphics[scale=0.35]{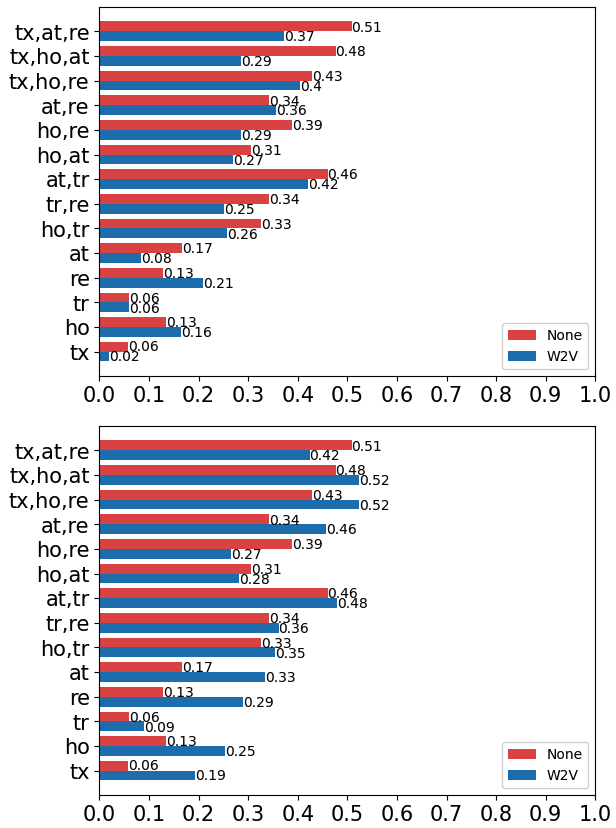} 
\vspace{-3mm}
\caption{Error rates per MultiWOZ domain category using the Word2Vec DA method and having \textbf{25\%} of the training data available. The upper and lower figures depict error rates when generating a single (\textbf{x2}) and four (\textbf{x5}) synthetic dialogs per instance respectively.}
\label{figures/error_woz_w2v.png}
\vspace{-5mm}
\end{center}
\end{figure}

In MultiWOZ dialogs the domain may change from utterance to utterance making the benchmark much more challenging, compared to KVRET. Using DA approaches have proven beneficial in terms of the overall \textit{Score}, but this is not always the case with the underlying metrics \textit{Success} and \textit{BLEU} score that concern the generated response as seen in Figures \ref{tab:10_x2_augment} - \ref{tab:25_x5_augment_kvret}. To gain a better view of the effectiveness of the DA methods tested, we present an error rate per domain category based on \textit{Success}. To do so, we calculate the percentage of unsuccessful dialogs ($\textit{Success} = 0$) per dialog category in the test set. Consider for example the case of using 25\% data of the training set for DA leveraging the Word2Vec approach. A comparison of this approach when using x2 and x5 the amount of data is depicted in Figure \ref{figures/error_woz_w2v.png}. Additional comparisons for the case of PEGASUS and fragment rotations are presented in Appendix \ref{sec:appendix_c_minus_two}. In the figure, ``tx'', ``at'', ``re'', ``ho'' and ``tr'' refer to the \textit{taxi}, \textit{attraction}, \textit{restaurant}, \textit{hotel}, and \textit{train} domains respectively. Multiple domains in each category denote domain switches. It comes as no surprise that the more domain switches a dialog contains, the more errors the model makes. Using x5 the amount of training data seems to amplify this behaviour; an observation that could be attributed to the additional noise introduced mostly in terms of the synthetically generated responses. Thus, generating more than a single synthetic dialog (especially in the x5 case) may not prove as beneficial for all cases (e.g. when 25\% of the MultiWOZ dialogs are available). Overall, of the three best performing methods, Word2Vec and PEGASUS seem to perform similarly, making more errors is the same dialog categories (mostly the ones including the restaurant and hotel domains). On the other hand fragment rotation straggles more mostly with dialogs involving the attraction domain. One should keep in mind that, regardless of applying DA or not, many errors made by the model are also to be blamed to the automatic evaluation as during inference the user's utterance remains static and is always in accordance with the ground truth system response, thus potentially confusing the model. 

After careful, manual examination of system responses on a small sample of the test set across all experimental cases, the responses themselves are fluent and informative. Generally, it is very difficult to deduce which DA method was used to train the model that generated the response. The small differences in scores could be justified by this conclusion, especially when comparing two methods that perform DA at the same level (e.g. the sentence-level PEGASUS and fragment rotation  methods). Even when generating multiple synthetic dialogs per training instance (e.g. x5 DA level), the responses remain fluent even though some DA methods may alter the structure of the sentence (e.g. Figure \ref{fig:rotate_dialog}). Using fragment rotation to generate four synthetic dialogs is an exception to the above observation, as in few cases the resulting responses may also feature rotated fragments of the respective dependency trees. An example of a dialog and its corresponding response per DA method is depicted in Figure \ref{figures/dialog_generation.png} of Appendix \ref{sec:appendix_c}. Moreover, a comparison of the models' outputs when generating a single and four synthetic dialogs per training instance is depicted in Figure \ref{x5_dialog_generation.png}.


\section{Advice to ToDS Practitioners} \label{sec:advice}

Based on our experiments, we provide the following advice to ToDS practitioners. (1) Although DA methods are currently not widely used in ToDS (Section~\ref{sec:related}), they can lead to very substantial gains when annotated training examples are scarce, 
even when using pre-trained end-to-end models. However, data scarcity depends on the difficulty of the task (e.g., whether the dialogues are single- or multi-domain), it cannot be judged using a universal threshold of initial annotated training instances, and is best assessed using performance scores. (2) Although all DA methods considered led to substantial improvements, dialog-state DA methods (at least the ones we considered) should be the last choices, as they led to a smaller performance boost and they also require dialog-specific annotations (e.g., dialog states) that are difficult to obtain in practice. (3) Among the other DA methods considered, the best overall performers were Word2Vec-based word substitution (Section~\ref{sec:3_1_1_subsection}), rotating parts of the dependency tree (Section~\ref{sec:3_2_3_subsection}), and paraphrasing the utterances by invoking paraphrasing models or prompting LLMs (Sections~\ref{sec:3_2_2_subsection}, \ref{sec:3_2_4_subsection}). We recommend using any of these methods, provided that the resources they require are available; e.g., rotation requires a dependency parser for the target language, prompting requires an LLM supporting the target language and adds the cost 
of invoking an LLM.
(4) Generating more than two synthetic instances per original one is worth considering only when the original annotated training instances are very scarce (e.g., when using only 2\% of MultiWOZ's multi-domain training set). 


\section{Related Work} \label{sec:related}

DA has been applied to a wide range of tasks from Computer Vision \citep{37, 38} to Speech Processing \citep{50} and NLP \citep{30,56,47, 49}.

For NLP specifically, \citet{21} carried out one of the most extensive comparisons of DA methods, in the context of biomedical QA, including Word2Vec-based word replacement (but without our semantics preserving test), LM-based word replacement, and back-translation. Word2Vec-based replacement was the most effective of the three methods, with back-translation and LM-based word replacement having similar performance. 
 
\citet{28} and \citet{29} prompt off-the-self LLMs to produce synthetic data in cross-lingual commonsense reasoning and intent classification,  respectively. \citet{74} also leverage an LLM to produce synthetic data for intent classification, slot filling, and next action prediction in a zero-shot ToDS setting, showing improvements in performance. By contrast, 
\citet{59} reported that relying on LLMs to directly generate belief states and system actions was not beneficial, compared to a fine-tuned model. 

\citet{12} propose the MADA (Multi-Action Data Augmentation) framework that we modified in order to include both the system actions and responses as explained in Section~\ref{sec:3_3_2_subsection}. \citet{11} also capitalise on dialog states to produce a tree of possible dialogs (Section~\ref{sec:3_3_1_subsection}). Both of these approaches proved beneficial in our work. Similar to the approach of \citet{11}, \citet{72} also capitalise on a graph structure, linking dialog states across all dialogs. In a similar spirit, \citet{25} propose clustering dialog turns based on their dialog states. The extracted cluster structure forms the base for an augmentation approach similar to MADA. 
\citet{71, 23},
inspired by \citet{36}, introduce an NLP version of image cropping and rotation, based on permutations of sentence fragments (Section \ref{sec:3_2_3_subsection}). To improve performance on out-of-scope (OOS) dialog utterances (not represented in the training set, e.g., pertaining to new domains), \citet{73} augment their training set by mixing OOS utterances with the ones found in their training set. \citet{68} 
use Reinforcement Learning to find the optimal (in terms of context) text span replacements (instead of word replacements).
 
Paraphrasing as a form of DA was also used by \citet{24} and \citet{22} for slot-tagging (subpart of DST) and RG; they implemented their own paraphrasing model based on RNNs. Similarly, \citet{79} train a BART  model \citep{80} for paraphrasing in an end-to-end ToD setting. \citet{27}, also employ DA in a ToDS setting, by substituting masked tokens of the input using an LM (much as in Section~\ref{sec:3_1_1_subsection}). \citet{69} augment their dataset by replacing slot-values with randomly generated strings to simulate references to out-of-vocabulary (OOV) words, a simple yet effective approach for models that learn to copy words from input to output by incorporating the copy-mechanism \citep{76}. \citet{84} randomly change the letter order of slot-values to improve the robustness of their ToDS. Improving robustness is also the main goal of \citet{85}, who generate turns with replaced slot-values not encountered in the training 
set. 

The closest previous work is that of \citet{26}. They experiment with four DA methods in an end-to-end ToDS setting, including stop-word removal, paraphrasing, back-translation, synonym substitution (and their combination). We compare more (eight) DA methods, of three different types (word-level, sentence-level, dialog-level), while Quan and Xiong consider fewer (four) methods of the word-level and sentence-level types only. Furthermore, we also experiment with a more challenging dataset containing dialogs that may belong (the same dialog) in multiple domains (MultiWOZ), whereas Quan and Xiong use the single-domain Camrest676 \cite{6}, and (like us) KVRET, which covers three domains, but each dialog pertains to a single domain. Contrary to the work of Quan and Xiong, we also experiment with several sizes of initial annotated training set (e.g., 2\%, 10\%, 25\% in MultiWOZ) and different expansions (x2, x3, x5), and we also introduce a new few-shot cross domain setting (Section~\ref{sec:main_setup}).


\section{Conclusions and Future Work} \label{sec:conclusion}

We performed the largest to date (in terms of methods and types) empirical comparison of DA methods in the context of end-to-end ToDSs, using two pre-trained Transformer ToDSs (UBAR, GALAXY) and two datasets (MultiWOZ, KVRET). We varied both the size of the initial annotated training set (e.g., 2\%, 10\%, 25\% of MultiWOZ's training set) and the expansion factor (generating one, two, or four synthetic instances from each original one). We also introduced a new, more challenging few-shot cross-domain evaluation setting. We showed that substantial performance gains can be obtained with DA methods, even when using pre-trained end-to-end models, and we offered concrete advice to ToDS practitioners regarding when to use DA, which DA methods to prefer, and with which expansion factors (Section~\ref{sec:advice}).

In future work, we plan to explore combinations of DA methods, to check if the synthetic training instances they produce are complementary. We also plan to investigate how well DA methods perform in low-resource languages (e.g., languages for which LLMs or reliable dependency parsers may not be available). Finally, we aim to measure the costs (computational, monetary) of DA methods.

\section{Limitations} \label{sec:limit}
A major limitation of our work is based on the nature of the datasets. Most of our DA methods rely on dialog annotations to ensure that they preserve the original semantics of the dialogs (Subsection \ref{sec:3_1_1_subsection}) or to directly generate synthetic dialogs (Subsection \ref{sec:3_3_1_subsection}). Many ToD datasets lack such detailed annotation, as is the case with the second track of DSTC 11 \citep{87}.\footnote{The dataset can be found at \url{https://github.com/amazon-science/dstc11-track2-intent-induction}.} In real world scenarios such annotations are even harder to come by, even though there is a plethora of structured dialogs covering a wide range of topics. In such cases, DA approaches that rely on the annotations of dialogs are not guaranteed to perform as well or at all on their own, compared to the experiments presented.

Another limitation of our work is the lack of detailed tuning/finetuning, in favour of a wider range of DA methods. Taking LLM-based paraphrasing as an example (Subsection \ref{sec:3_2_4_subsection}), although we do experiment (up to a point) with different prompts during preliminary testing, a more detailed prompt tuning may have led to better results. In the same manner, for the case of PEGASUS paraphrasing (Subsection \ref{sec:3_2_2_subsection}) or RoBERTa-based word substitution (Subsection \ref{sec:3_1_2_subsection}), further specialised pre-training on other similar ToD datasets could have led to improved performance as shown by \citet{1511}. Similarly we avoid tuning the hyperparameters (e.g., learning rate, context window etc.) of the models used (UBAR and GALAXY). After all, the purpose of our work was not to achieve SOTA performance, but to compare DA methods and show their benefits under a common task (end-to-end).

Finally, we only take into account DA methods that can be applied regardless of the ToDS used. ToDSs such as the ones proposed by \citet{22} and \citet{12} consider online DA methods, that were not covered in our work. We recognise the advantages of such methods but opt not to include them as they are ToDS specific (they are build around a specific ToDS). Instead we experiment with offline DA methods that are architecture-free and have proven beneficial in ToD and NLP in general.   

\section{Ethical considerations} \label{sec:ethics}
Implementing the DA methods and training the ToDS proposed in the settings mentioned in Section \ref{sec:setup}, requires a lot of GPU processing (excluding dialog-level DA and fragment rotation). Leaving aside the cost that comes as a result of the methods' execution, such GPU usage also has environmental impact in the form of CO2 emissions. We hope to alleviate this effect by offering advice to ToDS practitioners (Section \ref{sec:advice}) and guiding them in choosing the best fit (in terms of DA methods) for their task, in order to avoid such costly experimentation in the future.  

\section{Acknowledgements} \label{sec:acc}
This work was partially supported by Omilia Natural Language Solutions Ltd; the Research Center of the Athens University of Economics and Business (AUEB-RC); and project MIS 5154714 of the National Recovery and Resilience Plan Greece 2.0, funded by the European Union under the NextGenerationEU Program. Views and opinions expressed are however those of the author(s) only and do not necessarily reflect those of the European Union or European Commission-EU. Neither the European Union nor the granting authority can be held responsible for them.

\bibliography{acl2023}
\bibliographystyle{acl_natbib}

\appendix
\section*{Appendix}

\section{LLM prompting}
\label{sec:appendix}
We initially construct a simplistic template prompt, without giving any in-context example paraphrases, that has the following form : "\textit{Paraphrase the following sentence twice. Maintain as much information as possible intact. The sentence to paraphrase is : \{\} }". The "\textit{\{\}}" symbol denotes the placeholder for the utterance (not delexicalised) to be paraphrased. More elaborate templates were also tested (e.g. providing the LLM with the dialog history), such as \textit{"Your job is to augment the given utterance from a dialog between a user and an assistant. You will be given the dialog history as context. The user's utterances start with USER :, while the assistant's utterances with ASSISTANT :. Here is the dialog up to this point : \{1\} . Paraphrase the following sentence twice. Maintain as much information as possible intact. The sentence to paraphrase is : \{2\}"}, where "\textit{\{1\}}" and "\textit{\{2\}}" denote the placeholders for the dialog history and current utterance respectively. During preliminary testing such templates did not provide a significant improvement, and as a result we opted for the more cost-efficient approach.

\section{More Detailed Results}
\label{sec:appendix_b}
In this section we present the detailed results of all the experiments conducted, including the additional metrics that accompany each dataset. For MultiWOZ, we include the $\textit{Inform}$ and $\textit{Success}$ score along with $\textit{BLEU}$ score, as explained in Subsections \ref{sec:main_setup}. Similarly for the case of KVRET the $\textit{Match}$, $\textit{Success F1}$ and $\textit{BLEU}$ scores are reported. For each case, we first present the performance on the initial experiments where we select the 3 best performing DA methods followed by the results of these there methods in various other settings. From the results, we observe that the additional metrics reported tend to follow the trends of the main metric ($\textit{Score}$), especially $\textit{BLEU}$ and $\textit{Success}$.

\begin{table*}[p!]
\centering
\begin{tabular}{llllll}
\hline
\textbf{DA type} & \textbf{DA Method} & \textbf{Inform} & \textbf{Success} & \textbf{BLEU} & \textbf{Score} \\
\hline
\multirow{2}{4em}{No DA} & Full training set &
95.40 & 80.70 & 17.00 & 105.1 \\& 10\% training set & 76.37 & 53.73 & 10.83 & 75.88\\ 
  \hline
\multirow{2}{6em}{Word-level} & Word2Vec replace & 83.57 & 61.70 & 12.06 & 84.69\\ 
     & RoBERTa replace & 81.57 & 55.70 & 11.48 & 80.11\\ 
  \hline
  \multirow{4}{6em}{Sentence-level} & back-translation (FR) & 80.97 & 58.83 & 11.41 & 81.31\\   
   & PEGASUS paraphrase & 83.33 & 60.87 & 11.06 & 83.16\\   
   & Fragment rotation & 83.17  & 61.37 & 12.33 & 84.59\\
  & LLM paraphrase & 78.87  & 58.70 & 12.70 & 81.48\\
  \hline
  \multirow{2}{6em}{Dialog-level} & Dialog Tree & 78.83  & 57.23 &  11.90 & 79.94\\
  & Act-Response substitution & 79.33  & 58.90 & 11.66 & 80.78\\
  \hline
\end{tabular}
\caption{\label{tab:10_x2_augment}
\textbf{MultiWOZ metrics}, for all \textbf{8 DA methods}, using \textbf{10\%} of the original training dialogs,  producing a single synthetic dialog from each original one (\textbf{x2}).
}
\end{table*}

\begin{table*}[h!]
\centering
\begin{tabular}{lllll}
\hline
\textbf{DA Method} & \textbf{Inform} & \textbf{Success} & \textbf{BLEU} & \textbf{Score}\\
\hline
  10\% training set & 76.37 & 53.73 & 10.83 & 75.88\\ 
  \hline
  Word2Vec replace & 83.60 & 58.33 & 11.42 & 82.39\\ 
  \hline
  PEGASUS paraphrase & 82.37 & 58.83 & 10.50 & 81.10\\   
  \hline
  Fragment rotation & 80.87  & 58.93 & 11.52 & 81.42\\
  \hline 

\end{tabular}
\caption{\label{tab:10_x3_augment}
\textbf{MultiWOZ metrics}, for the \textbf{top 3 DA methods}, using \textbf{10\%} of the original training dialogs, producing two synthetic dialogs from each original one (\textbf{x3}).
}
\end{table*}

\begin{table*}[h!]
\centering
\begin{tabular}{lllll}
\hline
\textbf{DA Method} & \textbf{Inform} & \textbf{Success} & \textbf{BLEU} & \textbf{Score}\\
\hline
  10\% training set & 76.37 & 53.73 & 10.83 & 75.88\\ 
  \hline
  Word2Vec replace & 83.67 & 54.67 & 10.45 & 79.62\\ 
  \hline
  PEGASUS paraphrase & 82.93 & 58.07 & 9.73 & 80.23\\   
  \hline
  Fragment rotation & 81.73  & 57.47 & 10.10 & 79.70\\
  \hline 

\end{tabular}
\caption{\label{tab:10_x5_augment}
\textbf{MultiWOZ metrics}, for the \textbf{top 3 DA methods}, using \textbf{10\%} of the original training dialogs, producing four synthetic dialogs from each original one (\textbf{x5}).
}
\end{table*}

\begin{table*}[h!]
\centering
\begin{tabular}{lllll}
\hline
\textbf{DA Method } & \textbf{Inform} & \textbf{Success} & \textbf{BLEU} & \textbf{Score}\\
\hline
  25\% training set & 88.00 & 69.73 & 13.65 & 92.52\\ 
  \hline
  Word2Vec replace & 89.42   & 70.40 & 13.51 & 93.41\\ 
  \hline
  PEGASUS paraphrase & 88.90   &  69.27   & 13.08     & 92.17\\   
  \hline
  Fragment rotation & 87.57    & 69.93 & 13.91 & 92.66\\
  \hline 

\end{tabular}

\caption{\label{tab:25_x2_augment}
\textbf{MultiWOZ metrics}, for the \textbf{top 3 DA methods}, using \textbf{25\%} of the original training dialogs, producing a single synthetic dialog from each original one (\textbf{x2}).
}
\end{table*}

\begin{table*}[h!]
\centering
\begin{tabular}{lllll}
\hline
\textbf{DA Method} & \textbf{Inform} & \textbf{Success} & \textbf{BLEU} & \textbf{Score}\\
\hline
  25\% training set & 88.00 & 69.73 & 13.65 & 92.52\\ 
  \hline
  Word2Vec replace & 89.77 & 70.33 & 13.13     & 93.18\\ 
  \hline
  PEGASUS paraphrase & 89.00  & 71.03   & 12.37     & 92.38\\   
  \hline
  Fragment rotation & 88.17    & 69.60 & 13.18 & 92.06\\
  \hline 
\end{tabular}
\caption{\label{tab:25_x3_augment}
\textbf{MultiWOZ metrics}, for the \textbf{top 3 DA methods}, using \textbf{25\%} of the original training dialogs, producing two synthetic dialogs from each original one (\textbf{x3}).
}
\end{table*}

\begin{table*}[h!]
\centering
\begin{tabular}{lllll}
\hline
\textbf{DA Method} & \textbf{Inform} & \textbf{Success} & \textbf{BLEU} & \textbf{Score}\\
\hline
  25\% training set & 88.00 & 69.73 & 13.65 & 92.52\\ 
  \hline
  Word2Vec replace & 88.73 & 66.97 & 12.78 & 90.63\\ 
  \hline
  PEGASUS paraphrase & 88.20  & 66.13   & 10.92 & 88.08\\   
  \hline
  Fragment rotation & 88.37    & 67.63 & 11.67 & 89.67\\
  \hline 
\end{tabular}
\caption{\label{tab:25_x5_augment}
\textbf{MultiWOZ metrics}, for the \textbf{top 3 DA methods}, using \textbf{25\%} of the original training dialogs, producing four synthetic dialogs from each original one (\textbf{x5}).
}
\end{table*}

\begin{table*}[h!]
\centering
\begin{tabular}{lllll}
\hline
\textbf{DA Method} & \textbf{Inform} & \textbf{Success} & \textbf{BLEU} & \textbf{Score}\\
\hline
  2\% training set & 37.20 & 15.60 & 5.62  & 32.02\\ 
  \hline
  Word2Vec replace & 57.87 & 7.13 & 1.95  &  34.45\\ 
  \hline
  PEGASUS paraphrase & 73.33  & 14.20   & 2.82 & 46.59\\   
  \hline
  Fragment rotation & 62.40 & 12.10  & 3.23 &  40.53\\
  \hline 
\end{tabular}
\caption{\label{tab:2_x2_augment}
\textbf{MultiWOZ metrics}, for the \textbf{top 3 DA methods}, using \textbf{2\%} of the original training dialogs, producing a single synthetic dialog from each original one (\textbf{x2}).
}
\end{table*}

\begin{table*}[h!]
\centering
\begin{tabular}{lllll}
\hline
\textbf{DA Method} & \textbf{Inform} & \textbf{Success} & \textbf{BLEU} & \textbf{Score}\\
\hline
  2\% training set & 37.20 & 15.60 & 5.62  & 32.02\\ 
  \hline
  Word2Vec replace & 73.93 & 29.93 & 6.27  & 58.20\\ 
  \hline
  PEGASUS paraphrase & 73.50  & 34.50     &7.13   & 61.13\\   
  \hline
  Fragment rotation & 70.37 & 32.43  & 7.14  & 58.54\\
  \hline 
\end{tabular}
\caption{\label{tab:2_x3_augment}
\textbf{MultiWOZ metrics}, for the \textbf{top 3 DA methods}, using \textbf{2\%} of the original training dialogs, producing two synthetic dialogs from each original one (\textbf{x3}).
}
\end{table*}

\begin{table*}[h!]
\centering
\begin{tabular}{lllll}
\hline
\textbf{DA Method} & \textbf{Inform} & \textbf{Success} & \textbf{BLEU} & \textbf{Score}\\
\hline
  2\% training set &  37.20 & 15.60 & 5.62  & 32.02\\ 
  \hline
  Word2Vec replace & 75.17 & 37.03  &7.38   & 63.48\\ 
  \hline
  PEGASUS paraphrase & 75.50  & 41.70   & 7.60  & 66.20\\   
  \hline
  Fragment rotation & 71.83 & 39.03 & 7.25   & 62.68\\
  \hline 
\end{tabular}
\caption{\label{tab:2_x5_augment}
\textbf{MultiWOZ metrics}, for the \textbf{top 3 DA methods}, using \textbf{2\%} of the original training dialogs, producing four synthetic dialogs from each original one (\textbf{x5}).
}
\end{table*}

\begin{table*}[h!]
\centering
\begin{tabular}{lllll}
\hline
\textbf{DA Method} & \textbf{Inform} & \textbf{Success} & \textbf{BLEU} & \textbf{Score}\\
\hline
  10\% training set &  28.37 & 13.72 & 8.61 & 29.65\\ 
  \hline
  Word2Vec replace & 33.16 & 17.17 & 8.63  & 33.80\\ 
  \hline
  PEGASUS paraphrase & 38.30 & 19.53 & 8.64 & 37.56\\   
  \hline
  Fragment rotation & 33.25  & 17.93 & 8.90  & 34.49\\
  \hline 
\end{tabular}
\caption{\label{tab:10_x5_attr}
\textbf{MultiWOZ metrics}, for the \textbf{top 3 DA methods}, using \textbf{10\%} of the original training dialogs, producing four synthetic dialogs from each original one (\textbf{x5}) and leaving out the \textbf{attraction} domain.
}
\end{table*}

\begin{table*}[h!]
\centering
\begin{tabular}{lllll}
\hline
\textbf{DA Method} & \textbf{Inform} & \textbf{Success} & \textbf{BLEU} & \textbf{Score}\\
\hline
  10\% training set & 32.15  & 17.43  & 7.62 & 32.41 \\ 
  \hline
  Word2Vec replace & 36.21  & 22.17 & 7.77 & 36.95 \\ 
  \hline
  PEGASUS paraphrase & 40.69  & 26.06 & 7.97 & 41.35 \\   
  \hline
  Fragment rotation & 37.06  & 24.62 & 8.26 & 39.10\\
  \hline 
\end{tabular}
\caption{\label{tab:10_x5_hotel}
\textbf{MultiWOZ metrics}, for the \textbf{top 3 DA methods}, using \textbf{10\%} of the original training dialogs, producing four synthetic dialogs from each original one (\textbf{x5}) and leaving out the \textbf{hotel} domain.
}
\end{table*}

\begin{table*}[h!]
\centering
\begin{tabular}{lllll}
\hline
\textbf{DA Method} & \textbf{Inform} & \textbf{Success} & \textbf{BLEU} & \textbf{Score}\\
\hline
  10\% training set &  28.37  & 13.65  & 7.17  & 28.19\\ 
  \hline
  Word2Vec replace & 33.41 & 16.55  & 7.78 & 32.76\\ 
  \hline
  PEGASUS paraphrase & 34.55 & 19.60  & 8.51 & 35.59\\   
  \hline
  Fragment rotation & 37.61  & 22.20  & 8.55  & 38.45\\
  \hline 
\end{tabular}
\caption{\label{tab:10_x5_restaurant}
\textbf{MultiWOZ metrics}, for the \textbf{top 3 DA methods}, using \textbf{10\%} of the original training dialogs, producing four synthetic dialogs from each original one (\textbf{x5}) and leaving out the \textbf{restaurant} domain.
}
\end{table*}

\begin{table*}[h!]
\centering
\begin{tabular}{llllll}
\hline
\textbf{DA Type} & \textbf{DA Method} & \textbf{Match} & \textbf{Success F1} & \textbf{BLEU} & \textbf{Score} \\
\hline
\multirow{2}{4em}{No DA} & Full training set &
81.58 & 81.67 & 20.86 & 102.48 \\& 5\% training set & 58.68 & 64.99 & 11.44 & 73.27\\ 
  \hline
\multirow{2}{6em}{Word-level} & Word2Vec replace & 61.75 & 73.62 & 13.76 & 81.44\\ 
     & RoBERTa replace & 61.05 & 73.75 & 13.27 & 80.67\\ 
  \hline
  \multirow{4}{6em}{Sentence-level} & back-translation (FR) & 60.00 & 72.39 & 11.64 & 77.83\\   
   & PEGASUS paraphrase & 56.49 & 69.81 & 12.63 & 75.78\\   
   & Fragment rotation & 64.04  & 73.71 & 13.89 & 82.76\\
  & LLM paraphrase & 64.74  & 72.57 & 13.39 & 82.04\\
  \hline
\end{tabular}
\caption{\label{tab:5_x2_augment_kvret}
\textbf{KVRET metrics}, for all \textbf{6 DA methods}, using \textbf{5\%} of the original training dialogs, producing a single synthetic dialog from each original one (\textbf{x2}).
}
\end{table*}

\begin{table*}[h!]
\centering
\begin{tabular}{lllll}
\hline
\textbf{DA Method} & \textbf{Match} & \textbf{Success F1} & \textbf{BLEU} & \textbf{Score}\\
\hline
  5\% training set & 58.68 & 64.99 & 11.44 & 73.27\\ 
  \hline
  Word2Vec replace & 54.74  & 75.41 & 12.83 & 77.91  \\ 
  \hline
Fragment rotation & 56.05  & 74.98 & 11.93 & 77.44\\
  \hline 
  LLM paraphrase & 57.89  & 71.87 & 10.82  & 75.70 \\   
  \hline

\end{tabular}
\caption{\label{tab:5_x3_augment_kvret}
\textbf{KVRET metrics}, for the \textbf{top 3 DA methods}, using \textbf{5\%} of the original training dialogs, producing two synthetic dialogs from each original one (\textbf{x3}).
}
\end{table*}

\begin{table*}[h!]
\centering
\begin{tabular}{lllll}
\hline
\textbf{DA Method} & \textbf{Match} & \textbf{Success F1} & \textbf{BLEU} & \textbf{Score}\\
\hline
  5\% training set & 58.68 & 64.99 & 11.44 & 73.27\\ 
  \hline
  Word2Vec replace & 56.05  & 74.25 & 9.58 & 74.73  \\ 
  \hline
  Fragment rotation & 57.11  & 67.00 & 10.37 & 72.42\\
    \hline 
  LLM paraphrase & 53.16  & 74.10 & 11.35 & 74.97 \\   
  \hline 

\end{tabular}
\caption{\label{tab:5_x5_augment_kvret}
\textbf{KVRET metrics}, for the \textbf{top 3 DA methods}, using \textbf{5\%} of the original training dialogs, producing four synthetic dialogs from each original one (\textbf{x5}).
}
\end{table*}

\begin{table*}[h!]
\centering
\begin{tabular}{lllll}
\hline
\textbf{DA Method} & \textbf{Match} & \textbf{Success F1} & \textbf{BLEU} & \textbf{Score}\\
\hline
  10\% training set & 74.21 & 76.69 & 16.70 & 92.16\\ 
  \hline
  Word2Vec replace & 68.42 & 79.00 & 15.84 & 89.55  \\ 
  \hline
  Fragment rotation & 73.94  & 78.28 & 15.40 & 91.51\\
    \hline 
  LLM paraphrase & 74.47  & 77.42 & 15.96 & 91.91\\   
  \hline
\end{tabular}
\caption{\label{tab:10_x2_augment_kvret}
\textbf{KVRET metrics}, for the \textbf{top 3 DA methods}, using \textbf{10\%} of the original training dialogs, producing a single synthetic dialog from each original one (\textbf{x2}).
}
\end{table*}

\begin{table*}[h!]
\centering
\begin{tabular}{lllll}
\hline
\textbf{DA Method} & \textbf{Match} & \textbf{Success F1} & \textbf{BLEU} & \textbf{Score}\\
\hline
  10\% training set & 74.21 & 76.69 & 16.70 & 92.16\\ 
  \hline
  Word2Vec replace & 71.84  & 76.10 & 13.66 & 87.63  \\ 
  \hline
  Fragment rotation & 72.90  & 77.93 & 13.82 & 89.22\\
    \hline 
  LLM paraphrase & 73.94  & 79.48 & 13.81 & 90.53\\   
  \hline
\end{tabular}
\caption{\label{tab:10_x3_augment_kvret}
\textbf{KVRET metrics}, for the \textbf{top 3 DA methods}, using \textbf{10\%} of the original training dialogs, producing two synthetic dialogs from each original one (\textbf{x3}).
}
\end{table*}

\begin{table*}[h!]
\centering
\begin{tabular}{lllll}
\hline
\textbf{DA Method} & \textbf{Match} & \textbf{Success F1} & \textbf{BLEU} & \textbf{Score}\\
\hline
  10\% training set & 74.21 & 76.69 & 16.70 & 92.16\\ 
  \hline
  Word2Vec replace & 69.73   & 77.16 & 13.37 & 86.81  \\ 
  \hline
  Fragment rotation & 71.58  & 77.17 & 13.79 & 88.16  \\
    \hline 
  LLM paraphrase & 72.10  & 75.20  & 11.41 & 85.06 \\   
  \hline
\end{tabular}
\caption{\label{tab:10_x5_augment_kvret}
\textbf{KVRET metrics}, for the \textbf{top 3 DA methods}, using \textbf{10\%} of the original training dialogs, producing four synthetic dialogs from each original one (\textbf{x5}).
}
\end{table*}

\begin{table*}[h!]
\centering
\begin{tabular}{lllll}
\hline
\textbf{DA Method} & \textbf{Match} & \textbf{Success F1} & \textbf{BLEU} & \textbf{Score}\\
\hline
  25\% training set & 78.94 & 80.25 & 18.79 & 98.38\\ 
  \hline
  Word2Vec replace & 78.95  & 79.61 & 18.32 & 97.61   \\ 
  \hline
  Fragment rotation & 77.11  & 78.16 & 18.09 & 95.72\\
    \hline 
  LLM paraphrase & 80.00  & 80.38 & 16.54 & 96.72\\   
  \hline
\end{tabular}
\caption{\label{tab:25_x2_augment_kvret}
\textbf{KVRET metrics}, for the \textbf{top 3 DA methods}, using \textbf{25\%} of the original training dialogs, producing a single synthetic dialog from each original one (\textbf{x2}).
}
\end{table*}

\begin{table*}[h!]
\centering
\begin{tabular}{lllll}
\hline
\textbf{DA Method} & \textbf{Match} & \textbf{Success F1} & \textbf{BLEU} & \textbf{Score}\\
\hline
  25\% training set & 78.94 & 80.25 & 18.79 & 98.38\\ 
  \hline
  Word2Vec replace & 76.31 & 80.63 & 17.75 & 96.22    \\ 
  \hline
  Fragment rotation & 78.42  & 79.77 & 16.95 & 96.04\\
    \hline 
  LLM paraphrase & 78.69 & 78.94 & 16.34 & 95.15 \\   
  \hline
\end{tabular}
\caption{\label{tab:25_x3_augment_kvret}
\textbf{KVRET metrics}, for the \textbf{top 3 DA methods}, using \textbf{25\%} of the original training dialogs, producing two synthetic dialogs from each original one (\textbf{x3}).
}
\end{table*}

\begin{table*}[h!]
\centering
\begin{tabular}{lllll}
\hline
\textbf{DA Method} & \textbf{Match} & \textbf{Success F1} & \textbf{BLEU} & \textbf{Score}\\
\hline
  25\% training set & 78.94 & 80.25 & 18.79 & 98.38\\ 
  \hline
  Word2Vec replace &  75.53  & 78.14 & 14.91  & 91.75   \\ 
  \hline
  Fragment rotation & 76.58  & 78.68 & 15.19 & 92.81\\
    \hline 
  LLM paraphrase & 80.53  & 79.22 & 13.84 & 93.72 \\   
  \hline
\end{tabular}
\caption{\label{tab:25_x5_augment_kvret}
\textbf{KVRET metrics}, for the \textbf{top 3 DA methods}, using \textbf{25\%} of the original training dialogs, producing four synthetic dialogs from each original one (\textbf{x5}).
}
\end{table*}

\section{Additional error comparisons}
\label{sec:appendix_c_minus_two}

\begin{figure}[!ht]
\begin{center}
\includegraphics[scale=0.32]{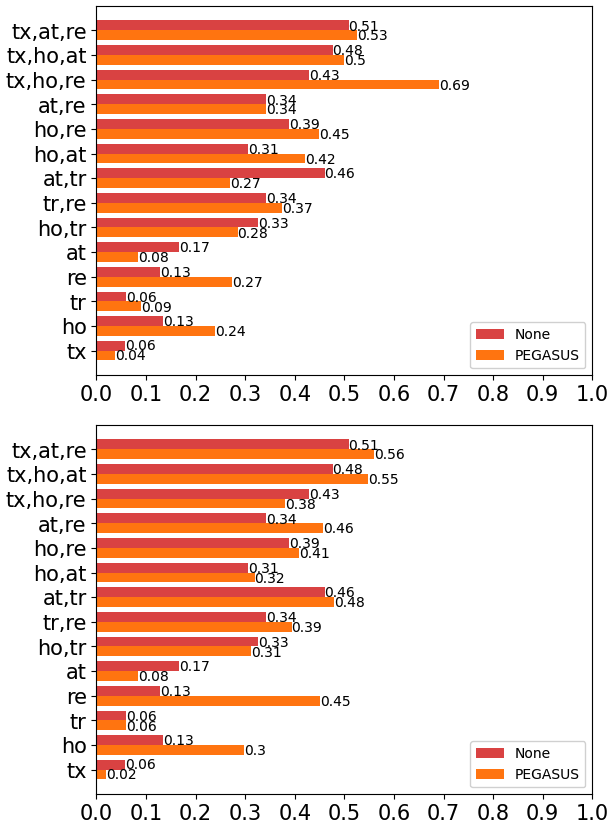} 
\vspace{-3mm}
\caption{Error rates per MultiWOZ domain category using the PEGASUS DA method and having \textbf{25\%} of the training data available. The upper and lower figure depict error rates when generating a single (\textbf{x2}) and four (\textbf{x5}) synthetic dialogs per instance respectively.}
\label{figures/error_woz_pegasus.png}
\vspace{-5mm}
\end{center}
\end{figure}

\begin{figure}[!ht]
\begin{center}
\includegraphics[scale=0.32]{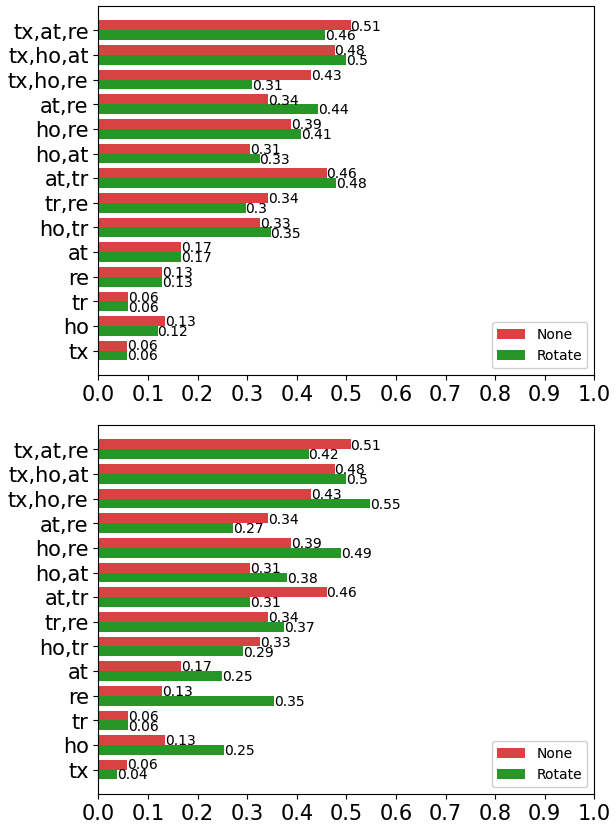} 
\vspace{-3mm}
\caption{Error rates per MultiWOZ domain category using the fragment rotation DA method and having \textbf{25\%} of the training data available. The upper and lower figure depict error rates when generating a single (\textbf{x2}) and four (\textbf{x5}) synthetic dialogs per instance respectively.}
\label{figures/error_woz_rotate.png}
\vspace{-5mm}
\end{center}
\end{figure}
We follow the same procedure of Section \ref{sec:analysis} for the cases of PEGASUS and fragment rotation. Once again, we count the number of unsuccessful dialogs $\textit{Success} = 0$ (i.e. the ToDS did not provide all the requested information or a valid recommendation), and divide by the number of dialogs belonging to each category, for the case where 25\% of the training dialogs of MultiWOZ was used. We focus on \textit{Success} as it features greater variance compared to \textit{BLEU} and \textit{Inform}, which generally remains unaffected. As was the case with Word2Vec, DA using PEGASUS or fragment rotation to generate more that one synthetic dialog reduces the overall \textit{Success}. Interestingly there are a few instances where using multiple synthetic dialogs managed to  increase the performance on specific, stand-alone domain categories such as ``tx, ho, re'' for PEGASUS and ``at, tr'' for fragment rotation, but such results may vary from case to case.

\section{Examples of synthetic and generated dialogs}
\label{sec:appendix_c}
\begin{figure}[H]
\begin{center}
\includegraphics[scale=0.45]{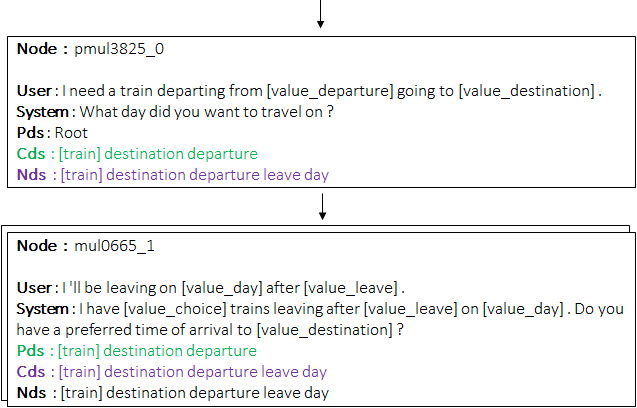} 
\caption{Two templates (from MultiWOZ dialogs $pmul3825$ and $mul0665$) are linked as they satisfy the dialog state conditions (shown in green and purple). }
\label{fig:d_tree}
\vspace{-5mm}
\end{center}
\end{figure}

In this section we present an example (Fig. \ref{fig:d_tree}) of the template matching performed by the dialog tree DA method (Subsection \ref{sec:3_3_1_subsection}). In this case the $\textit{Cds}$ of the first turn of $pmul3825$ matches the $\textit{Pds}$ of $mul0665$'s second turn. Similarly, $pmul3825$'s $\textit{Nds}$ matches $mul0665$'s $\textit{Cds}$, thus setting $mul0665$'s turn as the continuation of $pmul3825$. Repeating the process until reaching a ``Leaf'' node (turn), creates a new synthetic dialog.

\begin{figure*}
\begin{center}
\includegraphics[scale=0.6]{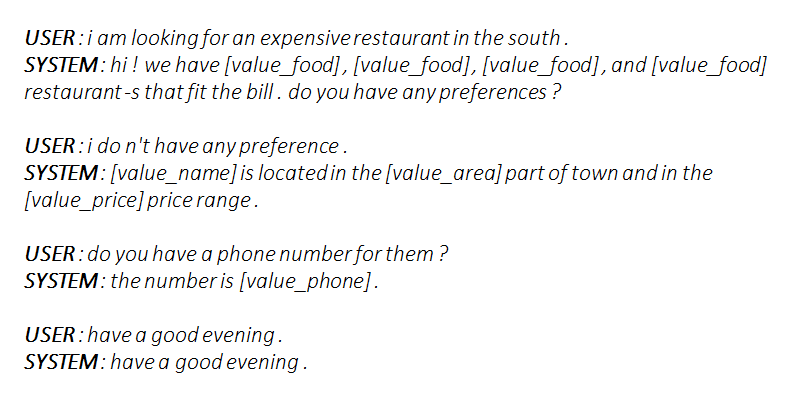} 
\caption{Dialog $woz20572$ of \textbf{MultiWOZ}, as used by \citet{17} for training UBAR, without any DA applied.}
\label{fig:original_dialog}
\end{center}
\end{figure*}

\begin{figure*}
\begin{center}
\includegraphics[scale=0.6]{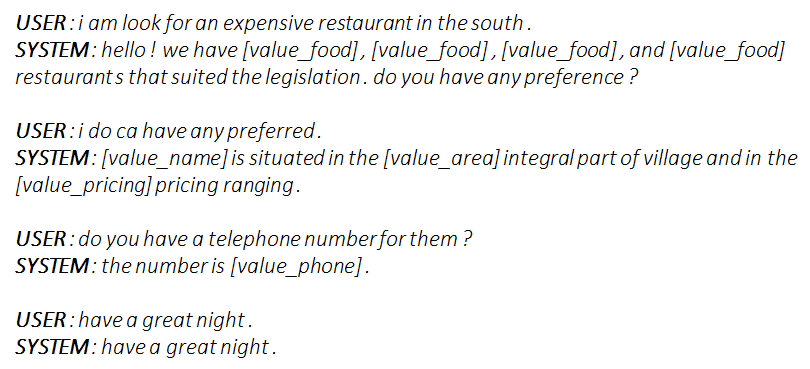} 
\caption{A synthetic dialog derived from $woz20572$ of \textbf{MultiWOZ} after applying our word substitution process via \textbf{Word2Vec}.}
\label{fig:w2v_dialog}
\end{center}
\end{figure*}

\begin{figure*}[h!]
\begin{center}
\includegraphics[scale=0.6]{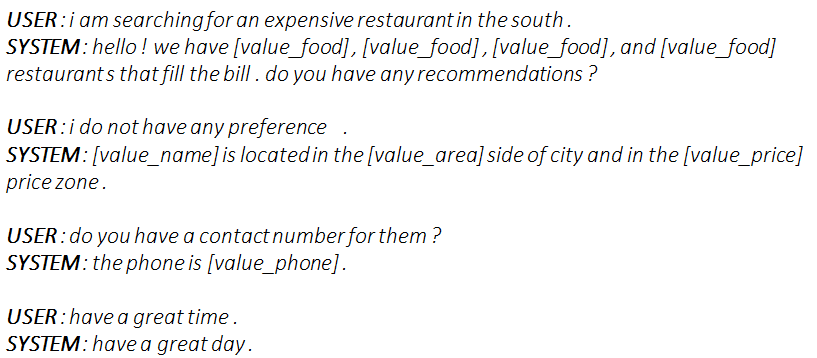} 
\caption{A synthetic dialog derived from $woz20572$ of \textbf{MultiWOZ} after applying our word substitution process using \textbf{RoBERTa}.}
\label{fig:roberta_dialog}
\end{center}
\end{figure*}

\begin{figure*}[h!]
\begin{center}
\includegraphics[scale=0.6]{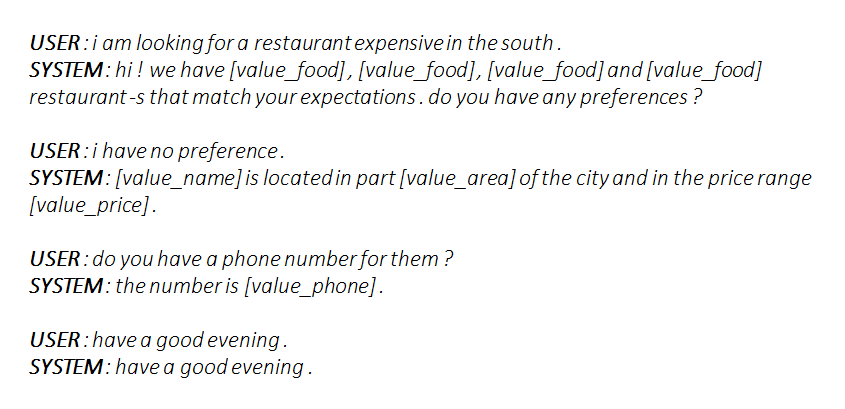} 
\caption{A synthetic dialog derived from $woz20572$ of \textbf{MultiWOZ} after applying \textbf{back-translation}. }
\label{fig:back_dialog}
\end{center}
\end{figure*}

\begin{figure*}[h!]
\begin{center}
\includegraphics[scale=0.6]{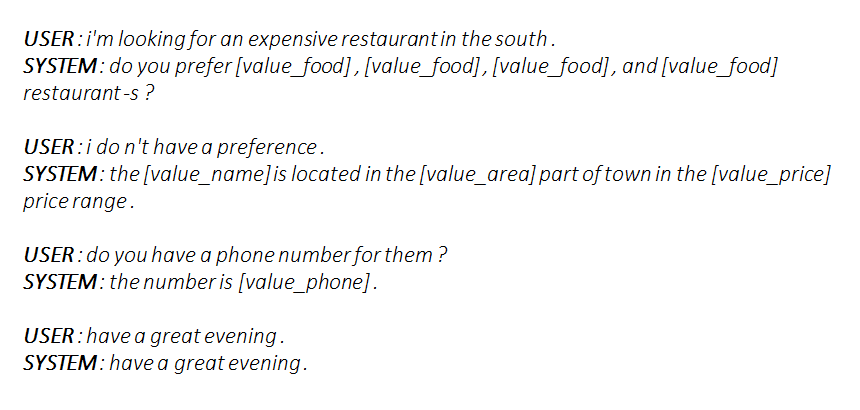} 
\caption{A synthetic dialog derived from $woz20572$ of \textbf{MultiWOZ} after paraphrasing using \textbf{PEGASUS}. }
\label{fig:pegasus_dialog}
\end{center}
\end{figure*}

\begin{figure*}[h!]
\begin{center}
\includegraphics[scale=0.6]{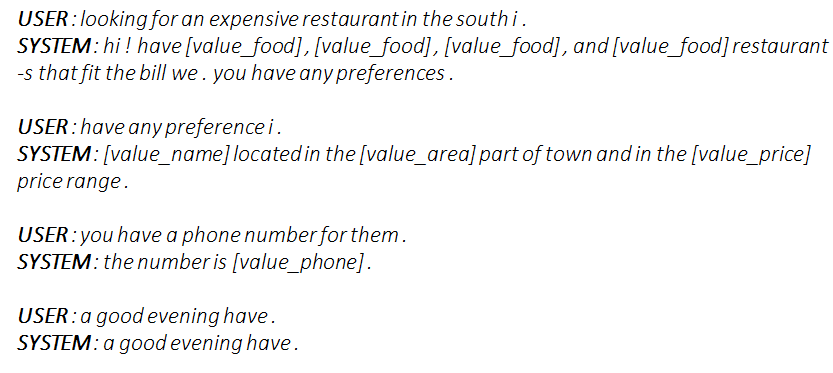} 
\caption{A synthetic dialog derived from $woz20572$ of \textbf{MultiWOZ} after applying \textbf{fragment rotation}. }
\label{fig:rotate_dialog}
\end{center}
\end{figure*}

\begin{figure*}[h!]
\begin{center}
\includegraphics[scale=0.6]{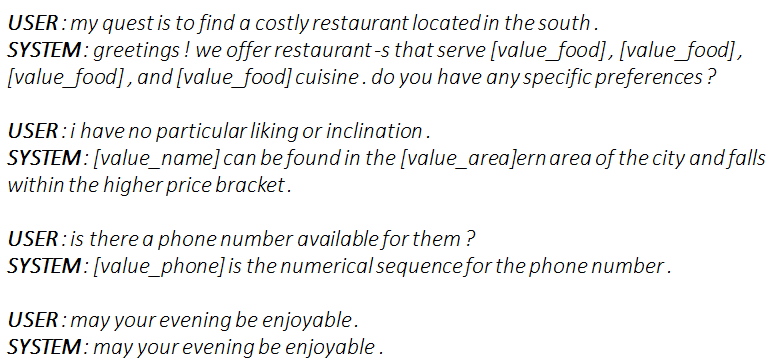} 
\caption{A synthetic dialog derived from $woz20572$ of \textbf{MultiWOZ} after paraphrasing using the GPT-3.5-turbo \textbf{LLM}. }
\label{fig:llm_dialog}
\end{center}
\end{figure*}

\begin{figure*}[h!]
\begin{center}
\includegraphics[scale=0.6]{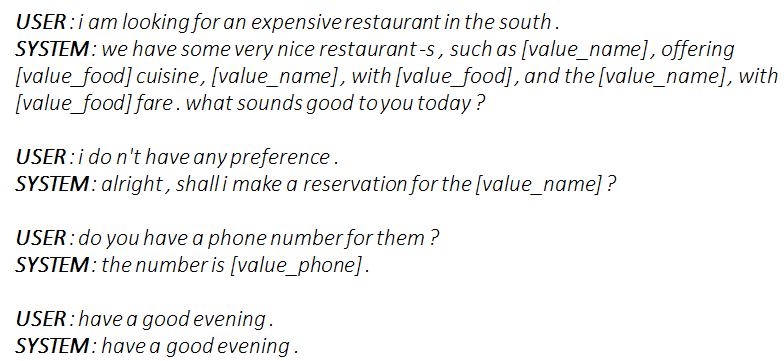} 
\caption{A synthetic dialog derived from $woz20572$ of \textbf{MultiWOZ} after applying our modified \textbf{MADA} approach.}
\label{fig:mada_dialog}
\end{center}
\end{figure*}

\begin{figure*}[h!]
\begin{center}
\includegraphics[scale=0.6]{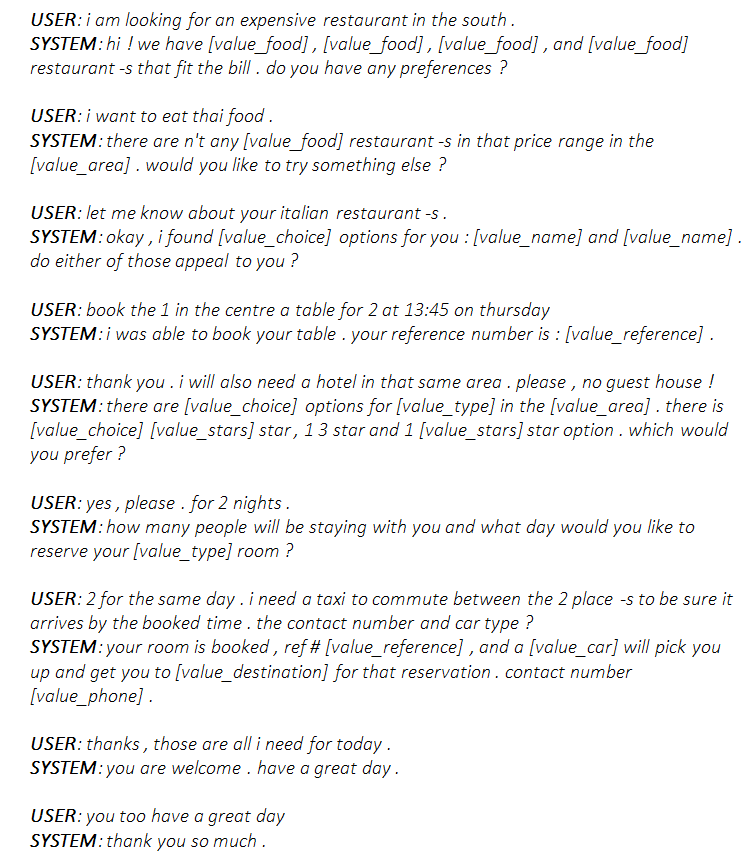} 
\caption{A synthetic dialog derived from the combination of $woz20572$ and $pmul4652$ of \textbf{MultiWOZ} using the \textbf{dialog tree} approach.}
\label{fig:tree_dialog}
\end{center}
\end{figure*}

Figures \ref{fig:w2v_dialog}-\ref{fig:tree_dialog} depict a single delexicalised dialog ($woz20572$) after applying each of the DA methods, for the case of MultiWOZ. An example of the same dialog, that UBAR is trained on, without any DA is depicted in Figure \ref{fig:original_dialog}.

\begin{figure*}[h!]
\begin{center}
\includegraphics[scale=0.19]{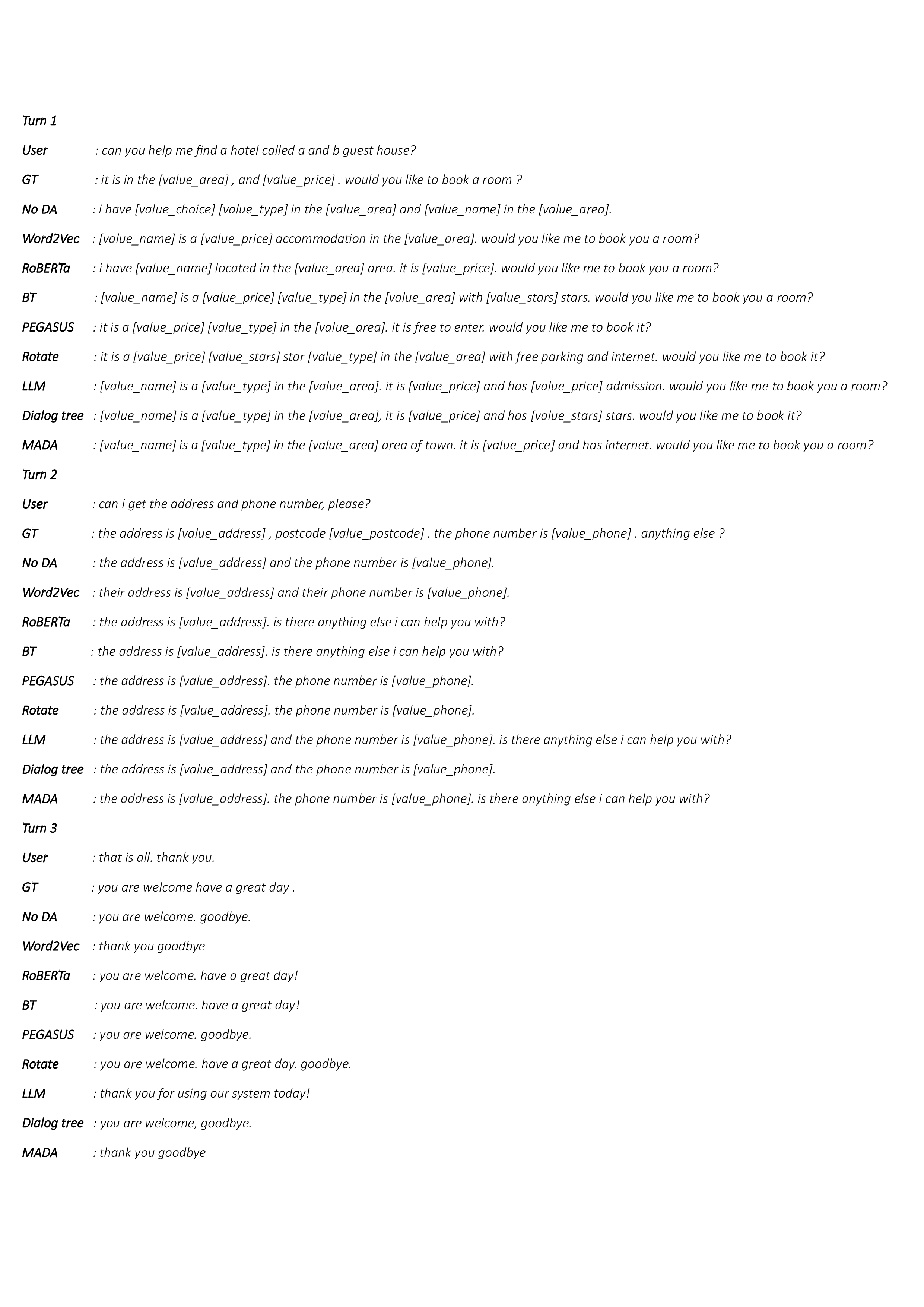} 
\caption{Dialog $sng01386$ of \textbf{MultiWOZ} along with the generated responses using all the DA methods described. The models where trained using 10\% of the training data and the augmentation level is \textbf{x2}. ``GT'' stands for ground truth while ``BT'' for back-translation.}
\label{figures/dialog_generation.png}
\end{center}
\end{figure*}

\begin{figure*}[h!]
\begin{center}
\includegraphics[scale=0.3]{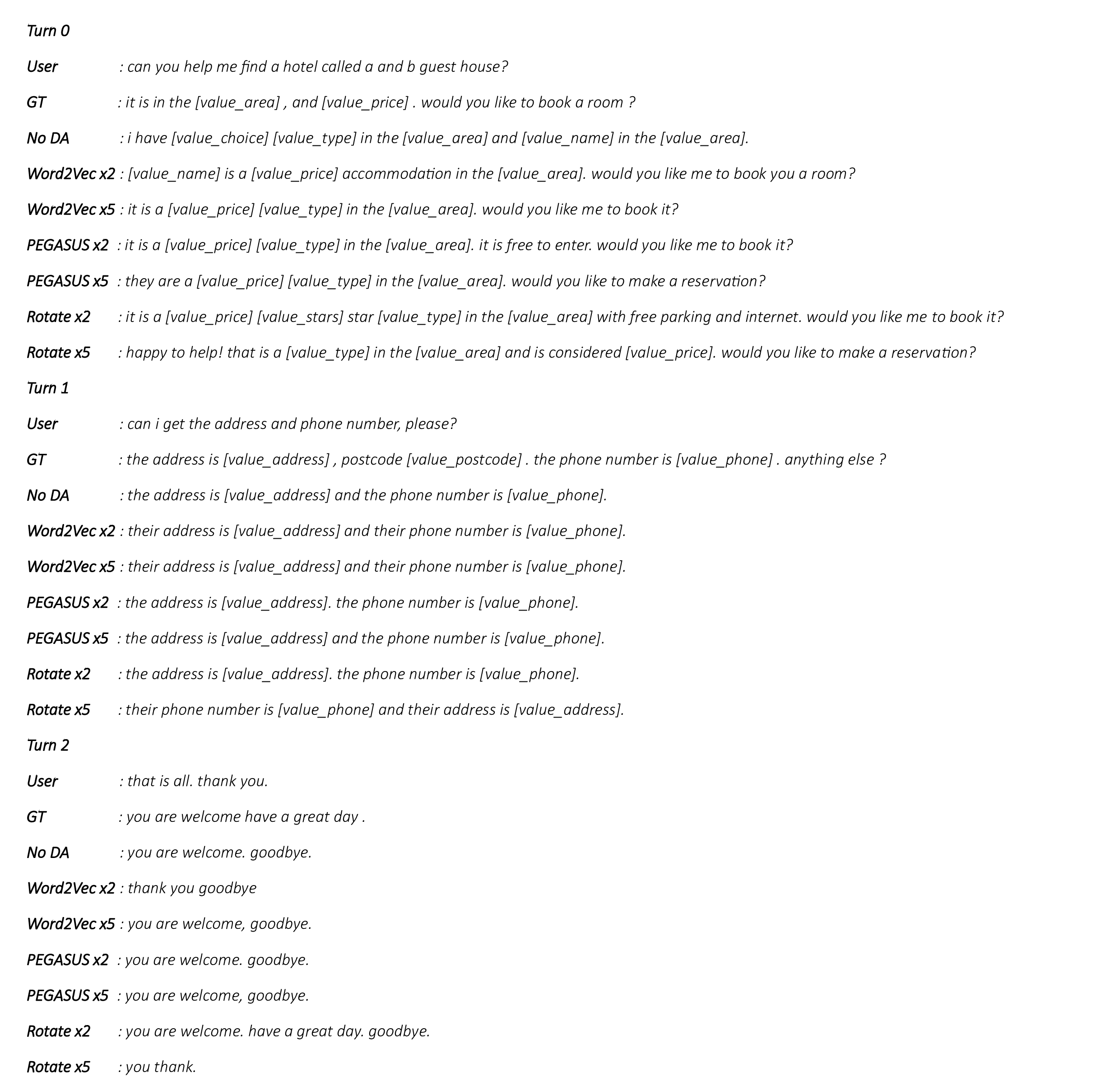} 
\caption{Dialog $sng01386$ of \textbf{MultiWOZ} along with the generated responses using the 3 best performing DA methods and 10\% of the available training dialogs. The figure depicts 2  responses per model (for x2 and x5 the amount of training data). ``GT'' stands for ground truth.}
\label{x5_dialog_generation.png}
\end{center}
\end{figure*}

\end{document}